\documentclass[10pt,twocolumn,letterpaper]{article}

\usepackage{cvpr}
\usepackage{times}
\usepackage{epsfig}
\usepackage{graphicx}
\usepackage{amsmath}
\usepackage{amssymb}

\usepackage{authblk}
\usepackage{xcolor}
\usepackage{adjustbox}
\usepackage{multirow}
\usepackage{caption}
\usepackage{enumitem}

\usepackage[pagebackref=true,breaklinks=true,letterpaper=true,colorlinks,bookmarks=false]{hyperref}

\makeatletter
\newcommand{\printfnsymbol}[1]{%
  \textsuperscript{\@fnsymbol{#1}}%
}
\makeatother

\cvprfinalcopy 

\def\cvprPaperID{7829} 

\ifcvprfinal\fi

\newcommand\blfootnote[1]{%
  \begingroup
  \renewcommand\thefootnote{}\footnote{#1}%
  \addtocounter{footnote}{-1}%
  \endgroup
}

\title{RefinedMPL: Refined Monocular PseudoLiDAR \\for 3D Object Detection in Autonomous Driving}

\author{Jean Marie Uwabeza Vianney$^*$}
\author{Shubhra Aich$^*$}
\author{Bingbing Liu$^\dagger$}
\affil{Noah's Ark Lab, Huawei Technologies, Markham, Canada\\
\texttt{\normalsize\{jeanm.vianney, shubhra.aich, liu.bingbing\}@huawei.com}
}

\begin{document}
\twocolumn[{%
\maketitle
\vspace{-5ex} 
\renewcommand\twocolumn[1][]{#1}%
\begin{center}
    \centering
\includegraphics[height=0.15\textwidth,angle=0]{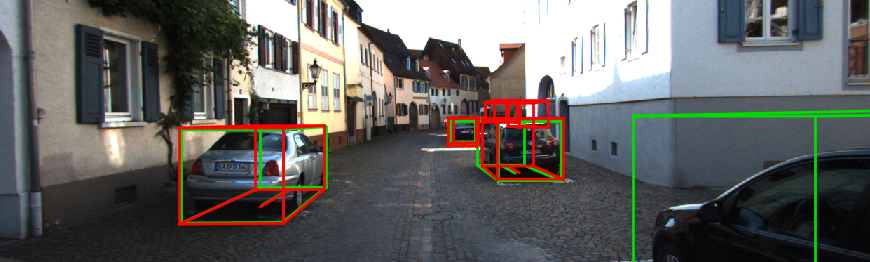} 
\includegraphics[height=0.15\textwidth,width=0.48\textwidth,angle=0]{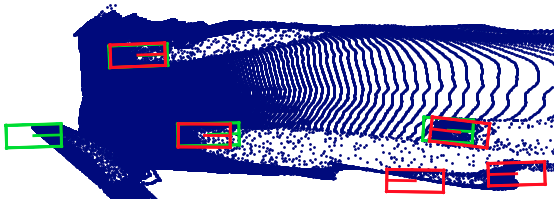} \\
\includegraphics[height=0.15\textwidth,angle=0]{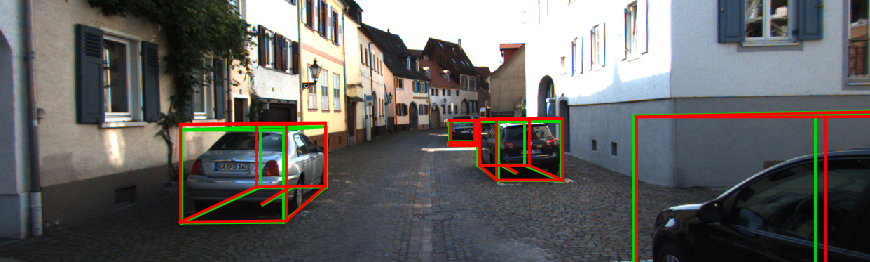} 
\includegraphics[height=0.15\textwidth,width=0.48\textwidth,angle=0]{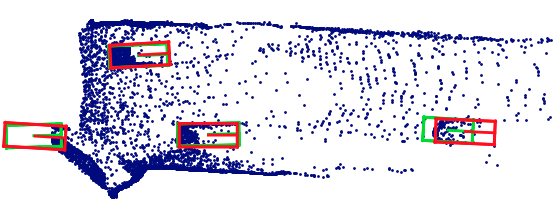} \\
\includegraphics[height=0.15\textwidth,angle=0]{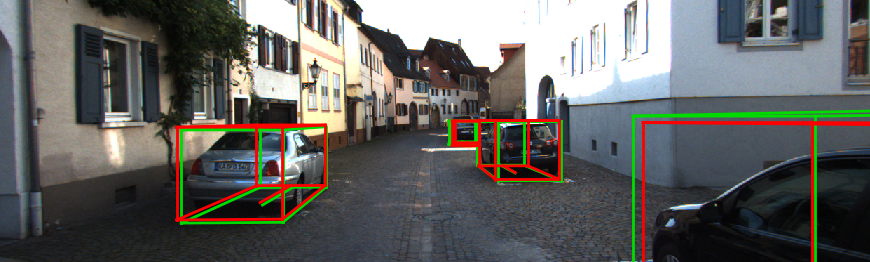} 
\includegraphics[height=0.15\textwidth,width=0.48\textwidth,angle=0]{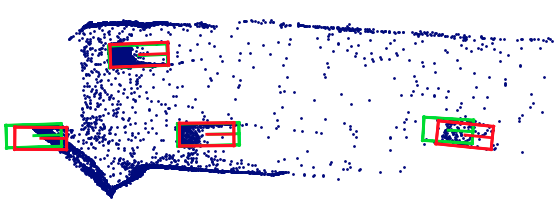} \\
\vspace{-1ex} 
    \captionof{figure}{Sample KITTI \cite{kitti3d} 3D detection results with PointRCNN \cite{point-rcnn} on the raw PseudoLiDAR (top), after unsupervised (middle) and supervised (bottom) sparsifications proposed in this paper. Right column contains cropped BEV representation of RGB (left) with superimposed ground truth and predictions shown in green and red, respectively. False positives caused by the overwhelming background density of PseudoLiDAR ($460K$ total points compared to $20K$ in Velodyne-64 LiDAR) are substantially mitigated after both sparsifications comprising only about $5\%$ of all the points. Best viewed in digital format.}
\label{fig:starter}
\end{center}%
}]

\blfootnote{\hspace{-4ex}$^*$equal contribution; $^\dagger$corresponding author}

\vspace{-2.5ex}
\begin{abstract}

\vspace{-1.5ex}

In this paper, we strive for solving the ambiguities arisen by the astoundingly high density of raw PseudoLiDAR for monocular 3D object detection for autonomous driving. Without much computational overhead, we propose a supervised and an unsupervised sparsification scheme of PseudoLiDAR prior to 3D detection. Both the strategies assist the standard 3D detector gain better performance over the raw PseudoLiDAR baseline using only $\sim 5\%$ of its points on the KITTI object detection benchmark, thus making our monocular framework and LiDAR-based counterparts computationally equivalent (Figure \ref{fig:starter}). Moreover, our architecture agnostic refinements provide state-of-the-art results on KITTI3D test set for ``Car" and ``Pedestrian" categories with $54\%$ relative improvement for ``Pedestrian". Finally, exploratory analysis is performed on the discrepancy between monocular and LiDAR-based 3D detection frameworks to guide future endeavours.

\end{abstract}

\vspace{-2.5ex}
\section{Introduction}
\label{intro}

It is prevailing for autonomous vehicles on the streets to employ LiDAR based solutions for robust 3D detection of on-road objects, such as cars, pedestrians, cyclists, etc. The accuracy of such a LiDAR detection pipeline depends heavily on the specificity of Laser scanning, i.e. the number of channels or laser beams, rotational speed, etc. This increasing level of granularity also increases the price of both the equipment and the vehicle notoriously. As an example, Velodyne HDL-64 LiDAR used in the initial prototype of Waymo self-driving cars\footnote{https://en.wikipedia.org/wiki/Waymo} costs about $75K$ \emph{USD}; thus increasing the expense of each vehicle by that huge amount. 

Although significant attempts have been made for 3D detection based on mono or stereo cameras \cite{NIPS2015_5644, 7932113, Li_2018_ECCV, Li_2019_CVPR, disentangle_kitti, deep-optics, m3d-rpn, mono-psr, roi-10d, fitting-degree, multi-fusion, deep-manta, multi-bin, mono3d-2016} to combat the high cost of LiDAR systems, almost all of these approaches are limited by the fact that they attempt to extend the 2D object detections from image space to 3D detections in object space. 

Recently, a few approaches \cite{pseudolidar, pseudolidar-copy, pseudolidar++} strive for solving the 3D detection problem exploiting the artificial point cloud generated from mono or stereo images than directly augmenting the 2D box information with geometric processing. Also, intriguing illustration is provided \cite{pseudolidar} regarding the besprinkling nature of 2D depth map under 2D convolution operator. This artificial point cloud is also known as \emph{PseudoLiDAR} due to the resemblance with LiDAR point clouds. 

PseudoLiDAR seems to bridge the gap between data modalities for camera based 3D detection. However, the raw PseudoLiDAR generated from single images contains a significant portion of background points, such as sidewalls, off-road objects, etc. In general, the 2D object detection algorithms based on deep learning \cite{faster-rcnn, mask-rcnn, yolo9000, ssd} can differentiate the candidate objects or foregrounds with substantially high accuracy compared to background things and stuff \cite{panoptic}, thanks to abundant contextual and texture features present in RGB images. The domain of 3D point clouds almost completely lacks such spatial cues (unless augmented or fused with RGB intensities \cite{pseudolidar-copy}), except for the reflectance (missing in PseudoLiDAR) and visible portion of the object envelopes. The prominence of background points in PseudoLiDAR along with the paucity of contextual information makes 3D detectors fail intriguingly (Figure \ref{fig:starter} top), otherwise doing a splendid job on the real LiDAR point clouds. Based upon this observation, we hypothesize that PseudoLiDAR need more attention before feeding it to the 3D detector, which is the stimulus to this work.

Our contribution in this paper is on the engineering side of PseudoLiDAR point cloud to facilitate the 3D detection task for the detection model. Following the recent literature \cite{point-rcnn, pointpillars, voxelnet, frustum-pointnet, pointnet++, pointnet}, from the architectural perspective, we argue that the LiDAR based 3D detection approaches are mature enough to localize the target objects (vehicles, pedestrians, cyclists) and obstacles with more or less satisfactory level. Hence, to uplift the monocular 3D detection performance, we need to focus on the discrepancy between PseudoLiDAR and actual LiDAR point clouds more closely. Recent works either produce the PseudoLiDAR of the whole scene \cite{pseudolidar, pseudolidar-copy, pseudolidar++} followed by processing with the detector network or transform the 2D scene information with 3D parameters for 3D bounding box regression \cite{NIPS2015_5644, 7932113, Li_2018_ECCV, Li_2019_CVPR, disentangle_kitti, deep-optics, m3d-rpn, mono-psr, roi-10d, fitting-degree, multi-fusion, deep-manta, multi-bin, mono3d-2016}. None of these approaches delve deep enough into the characteristics of the PseudoLiDAR point cloud itself. In this regard, this work is the first to cast light on the data engineering perspective of monocular 3D detection based on PseudoLiDAR point clouds. 

As the PseudoLiDAR point cloud is generated from the pixel-wise monocular depth map, it is much denser compared to the 64/128 channel LiDARs. Also, the 3D detectors perform pretty well with much sparse 64 channel point clouds. Therefore, the high density of PseudoLiDAR does not add much to the final performance of the mono 3D pipeline. Instead, careful reduction of the density of PseudoLiDAR would make the pipeline both computationally and performance-wise efficient, since most of the points in the high-density point clouds belong to the background as shown in Figure \ref{fig:starter}. This hypothesis is also proved on the KITTI validation set with thorough experimentation presented later. 

\begin{figure*}[!ht]
\centering
\includegraphics[width=\textwidth,angle=0]{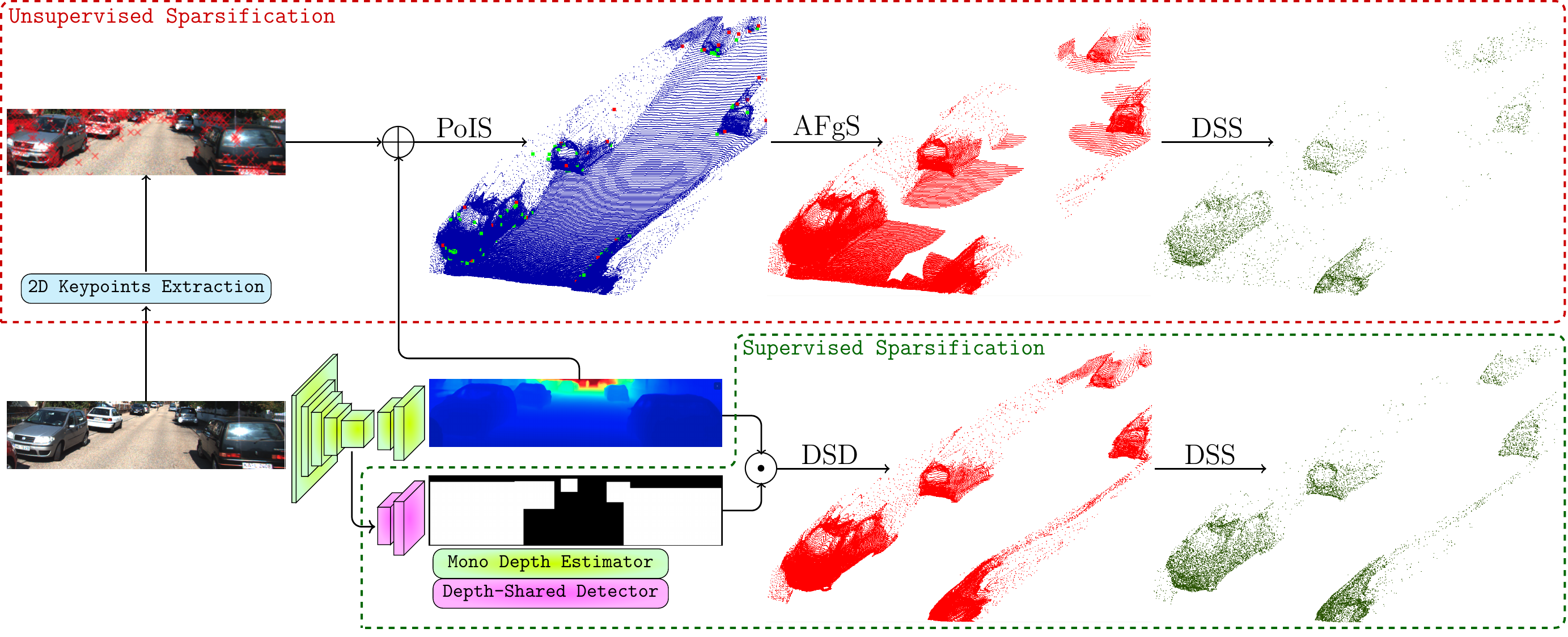}
\vspace{-3.5ex}
\caption{Illustration of unsupervised (top) and supervised (bottom) sparsification. \emph{PoIS $\equiv$ points of interest selection; AFgS $\equiv$ approximate foreground separation; DSS $\equiv$ distance stratified sampler; DSD $\equiv$ depth-shared 2D detector}. (Top) The keypoints before and after keypoint-sparsification are indicated by \emph{green} and \emph{red} in the raw PseudoLiDAR (\emph{blue}). 
Best viewed in digital format.}
\label{fig:visualization}
\end{figure*}

To this end, we propose a couple of complete preprocessing pipelines -- one unsupervised and another supervised, combining existing and novel practical schemes to generate refined PseudoLiDARs from mono depth maps (Figure \ref{fig:visualization}). The unsupervised proposition primarily comprises three basic modules as follows: 

\vspace{-1.ex}
\begin{itemize}[leftmargin=15pt]
    \small \item  \emph{Points of Interest Selection (PoIS)} to determine and sample interest points for foreground segmentation. \vspace{-1.7ex}
    \small \item  \emph{Approximate Foreground Separation (AFgS)} for search based foreground separation in 3D point space.  \vspace{-1.5ex}
    \small \item \emph{Distance Stratified Sampler (DSS)} to sample foreground points preserving ratio of object density. 
\end{itemize}

\vspace{-0.5ex}
\noindent And, the elements of the supervised maneuver are given by 
\vspace{-0.5ex}
\begin{itemize}[leftmargin=15pt]
    \small \item  \emph{Depth-Shared 2D Detector (DSD)} for supervised 2D segmentation/detection reusing mono depth features. \vspace{-1.7ex}
    \small \item \emph{Distance Stratified Sampler (DSS)} (same as unsupervised).
\end{itemize}

\noindent All of these constituents are described in detail in Section \ref{sec:approach}. Both of our schemes provide substantial performance improvement on the \textit{validation set} of the KITTI object detection benchmark with a dramatic reduction in computation cost compared to raw PseudoLiDARs. In addition, our supervised strategy provides SOTA results for \emph{``Car"} and \emph{``Pedestrian"} on KITTI3D \textit{test set} with $54 \%$ relative improvement on the \emph{``Pedestrian"} category. The unsupervised approach is preferred to the supervised one in case a large portion of the objects are unlabeled, which might make the training of the supervised approach comparatively difficult to converge. Finally, we put an effort to analyze the inherent limitations of the monocular approaches compared to their LiDAR counterparts due to the highly ill-posed nature of depth prediction.

\noindent Our contributions can be summarized as follows:
\vspace{-1ex}
\begin{itemize}[leftmargin=*]
    \item We show the infelicitous nature of the raw, dense PseudoLiDAR comprising a significantly high number of background points for 3D detection.\vspace{-1.5ex}
    \item To improve both the numerical and computational performance of 3D detectors on PseudoLiDARs, we propose an unsupervised and a supervised preprocessing schemes with existing and partially novel ideas. To our knowledge, this is the first work putting the accent on PseudoLiDAR post-orchestration. \vspace{-1.5ex}
    \item Extensive evaluation on the KITTI3D validation set demonstrates the superiority of both of our approaches over the baseline. In addition, we achieve SOTA performance on \emph{``Car"} and \emph{``Pedestrian"} categories on KITTI3D test set with $54 \%$ improvement over the recent literature. \vspace{-1.5ex}
    \item Interpretation of the discrepancy of performance between monocular and LiDAR-based 3D detection approaches are provided to usher future attempts to bridge this gap. \vspace{-1.5ex}
\end{itemize}

\section{Related Work}

In this section, we provide a detailed overview of the recent monocular 3D detection methods for autonomous driving. A short brief of the emergence of 3D detection architectures is also provided since we exert one of these networks for final detection in this paper.

\subsection{Monocular 3D Detection}

Most of the monocular 3D detection approaches somewhat attempt to enhance the 2D bounding box information with 3D geometric information for which incorporation of strong 2D detection priors appears to be inevitable. Chen \emph{et al.} \cite{mono3d-2016} exploit the heuristic of ground-plane proximity and multiple image features, including class and instance semantics, shape, context, and location priors to score the 3D detection proposals. Mousavian \emph{et al.} \cite{multi-bin} emphasize on the constraint that the projected 3D bounding boxes should fit tightly inside 2D detections in the image space. DeepMANTA \cite{deep-manta} solves the 3D detection problem by optimizing for multiple or many related tasks, i.e. 2D box regression, classification, part visibility, and template similarity, altogether. The final 3D pose is computed with the best-matched 3D template after non-max suppression. 

The fusion of mono image with multi-modal data, such as disparity or depth and point cloud generated from the image itself is first performed in \cite{multi-fusion} to facilitate 3D object detection. MonoSS3D \cite{mono-ss3d} mimics the principle of single-shot 2D detectors \cite{ssd, yolo9000} for 3D object detection with a 26D surrogate targets instead of 7. 

MonoPSR \cite{mono-psr} leverages 2D detection with concatenated features from both resized and cropped inputs together. Also, it learns residual depth with respect to the output of a pinhole camera model for better stability. Finally, instance reconstruction loss is computed in all three coordinate frames, namely object, camera, and pixel coordinates. 
ROI-10D \cite{roi-10d} regresses or lifts the 2D detections directly based on 2D ROI-Aligned feature and coordinate maps \cite{coord-conv}, and mono depth predictions.
Liu et al. \cite{fitting-degree} propose to solve the 3D detection problem by inferring 3D IoU between 3D proposals generated by Gaussian dense sampling and object using 2D information from the mono image.
MonoGRNet \cite{mono-grnet} learns instance-wise depth and 3D bounding box parameters in the image frame based on 2D detection results.

Deep Optics \cite{deep-optics} attempts to model the monocular depth estimation task as an optical-encoder and electronic-decoder system with the advantage of freeform lenses over conventional ones.
M3D-RPN \cite{m3d-rpn} simultaneously optimizes the 2D-3D detection parameters with a single region proposal network with iterative optimization for 3D rotations using 3D-2D projections.
MonoDIS \cite{disentangle_kitti} proposes disentangled training of the losses from heterogeneous sets of parameters, and signed IoU metric based on the rotation of predicted boxes. Also, the subtlety of zero recall on the KITTI metric \cite{kitti3d} is first addressed here.

The deficit of the 2D input representation for 3D detectors is extensively studied very recently \cite{pseudolidar, pseudolidar-copy, pseudolidar++}, one of which coined the term PseudoLiDAR \cite{pseudolidar}. Severe distortion or dispersion of the depth map occurs upon applying 2D convolution on it since the dense operators like 2D convolution are assumed to be working well in the regions of spatial proximity. Instead of using the depth information as an auxiliary channel or fusing it for 2D multi-tasking, utilizing it to reproject the image pixels into the 3D space, and processing the 3D points for further detection outplays the aforementioned mono 2D counterparts. 

Our supervised depth-shared 2D detection approach conveys similar spirit as in \cite{pseudolidar-copy}, where two separate 2D bounding box regression and depth prediction models are trained to generate object-specific point clouds followed by point cloud segmentation. However, we exploit the depth encoder for coarse localization of the candidate objects in the image, thus sharing a significant amount of computation in the preprocessing stage. Our ablation studies show that the depth-shared feature maps indeed assist in overall 3D detection performance. Additionally, 2D box scores are propagated to the output of the 3D detectors \cite{pseudolidar-copy}. Although such propagation from a separately trained model might be useful, we find it redundant as the classification and scoring capability of the 3D model is strong enough to differentiate between object categories. Furthermore, we argue that training the 3D detector to identify object categories directly acts as a good compelling force to detect more robust features for 3D box regression in the multi-tasking environment.

\subsection{3D Detection Architectures}

Although it is quite common to represent the unstructured, irregular 3D point cloud collected with LiDAR into regular bird's eye view (BEV) to exploit the full potential of 2D convolutional networks \cite{mv3d, avod, pixor, continuous-fusion, hdnet}, methods directly operating in the 3D point space \cite{frustum-pointnet, multiview-frustum-pointnet, point-rcnn, fast-point-rcnn} still tend to provide better performance. Therefore, we have chosen an architecture from the later family with a short brief of the recent advancements in this direction here.

PointNet family of architectures \cite{pointnet, pointnet++} learn pointwise features from the neighborhood of the chosen metric space. The vanilla Pointnet \cite{pointnet} is enhanced in PointNet++ \cite{pointnet++} by hierarchical feature extraction with iterative farthest point sampling (FPS) and ball query. For dense point labeling tasks, cluster features are propagated with weighted interpolation based on inverse distance.

VoxelNet \cite{voxelnet} partitions the 3D space into voxel grids followed by voxel feature extraction using PointNet. However, 3D convolution used on voxel features and granularity of the voxel grid transpire to be a huge bottleneck for efficiency. PointPillars \cite{pointpillars} encodes the LiDAR point cloud in pretty much the same way as VoxelNet, except that a single unit is considered along the $Z$-axis, which means discretization is done only on the $XY$-plane for significant speedup. 

PointRCNN \cite{point-rcnn} can be considered the first 3D-point based doppelganger of the popular RCNN family of architectures \cite{rcnn, fast-rcnn, faster-rcnn, mask-rcnn}, except that it uses discrete bins instead of anchors for the 3D proposal generation stage. Fast-PointRCNN \cite{fast-point-rcnn} proposes to voxelize the point space prior to region proposal generation to overcome the computational burden regarding huge number of points in the raw point cloud which inherently incorporates the trade-off between the granularity of the voxel grid and computational efficiency. Note that our approach effectively reduces the number of points by more than 95\% without sacrificing accuracy. Therefore, we have chosen PointRCNN as our 3D detector to leverage the full potential of the point space \cite{point-rcnn, fast-point-rcnn}.

\section{Our Approach}
\label{sec:approach}

Visual illustration of both unsupervised and supervised approaches are provided in Figure \ref{fig:visualization}. As already mentioned, both of our schemes can be considered a form of objective sparsification of dense PseudoLiDAR to mitigate its numerical and computational bottleneck. 

\subsection{Unsupervised Sparsification}
Our unsupervised proposal to sparsify the PseudoLiDAR point clouds is done in three steps: (1) interest point detection and sampling, (2) approximate foreground separation with nearest neighbor search, and (3) global sampling stratified by distance. 

\vspace{-1ex}
\subsubsection{Points of Interest Selection (PoIS)}

We select the candidate or interest points in two steps. First, the interest points are extracted with Laplacian of Gaussian (LoG) extrema detection \cite{szeliski} on the forward difference image with a predefined step-size. The forward difference image appears to be more robust compared to the raw intensity counterpart. The former leads to the keypoints around strong directional derivatives shifted or parameterized by the step-size of the forward difference image. 
Second, fewer candidate keypoints are selected using nearest neighbor (NN) clustering in the 3D space with a predefined search radius. This clustering operation incurs negligible overhead due to a much lower number of keypoints with the advantage of further sparsification during foreground separation later on. 

\vspace{-1ex}
\subsubsection{Approximate Foreground Separation (AFgS)}

We exploit the selected keypoints from the previous step for further sparsification over the global point clouds. Our approximate foreground region is defined by the second nearest neighbor query around the keypoints selected in the first step. However, such simpler heuristics based refinement does not guarantee foreground separation with perfect recall. We address this issue by adding a lower number of points from the background point sets as well in our final, preprocessed point clouds containing both foreground and background (lower) points, sampled by the distance-stratified sampler.

\vspace{-1ex}
\subsubsection{Distance-Stratified Sampler (DSS)}
\label{sec:dss}

PseudoLiDAR generated from a dense depth map contains a plethora of points for particular kinds of background objects (such as trees, bushes, sidewalls, etc.) as well as foreground objects. For example, the number of points in the whole scene PseudoLiDAR and Velodyne-64 LiDAR in Figure \ref{fig:visualization} are approximately $430K$ and $20K$, respectively.

Even the approximately separated foreground in PseudoLiDAR contains a large number of points compared to its LiDAR counterpart. In Figure \ref{fig:visualization}, the number of foreground points before DSS for unsupervised and supervised schemes are about $104K$ and $133K$ respectively, whereas the corresponding Velodyne-64 LiDAR contains $6.6K$ ($\sim 95\%$ less) foreground points. Moreover, a significant portion of these surplus points are in the nearby objects, which do not necessarily need these excess points for precise localization. The comparatively lower number of extra points in the faraway and/or partially occluded objects might be useful for their detection. 

Therefore, even a marginal reduction of the number of points would be highly beneficial from the computational perspective. A naive solution is to randomly sample from the point clouds universally. However, given that the comparatively much higher number of points from the larger instances that are situated closer to the camera, it runs out of the risk of completely wiping out smaller, faraway foreground objects significantly, resulting in severe degradation of the detection accuracy. 

Instead, we need a somewhat object-specific sampling in our scene that can be accomplished with the distance based sampling as the surrogacy. We argue that most of the hard-to-detect instances are indeed smaller, distant ones in the background. Hence, one of the most plausible ways to preserve a non-negligible number of points from those hard objects is to sample the points uniformly based on the range of the points. Since, in this process, the points are stratified according to their distance before sampling, we call this strategy distance-stratified sampler (DSS). 

Also, ablation study (Section \ref{sec:ablation}) shows that reducing the point density of foreground sub-regions with DSS does not hamper the mean APs. The reason behind such stability is hypothesized to be the coherent spatial structure present in the foreground 3D points preserved by DSS. 

\vspace{1ex}
\subsection{Supervised Sparsification}

Our supervised data engineering scheme involves depth-shared foreground separation followed by distance stratified sampling (Figure \ref{fig:visualization}). We only describe depth-shared detection here since DSS is the same as in Section \ref{sec:dss}.

\vspace{-2ex}
\subsubsection{Depth-Shared 2D Detector (DSD)}

We argue that PseudoLiDAR point clouds inherently contain more \emph{a priori} information from mono or stereo images than the LiDAR counterparts. The most obvious prior knowledge is the contextually rich foreground map including both nearby and faraway objects which is exploited in numerous image based 3D detection approaches \cite{NIPS2015_5644, 7932113, Li_2018_ECCV, Li_2019_CVPR, disentangle_kitti, deep-optics, m3d-rpn, mono-psr, roi-10d, fitting-degree, multi-fusion, deep-manta, multi-bin, mono3d-2016}. 

However, all these approaches incorporate a separate 2D detection or segmentation model bouncing up the computational complexity significantly obliterating the possible advantages of foreground-only postprocessing. We rejuvenate the idea of feature sharing to mitigate this computational bottleneck. In general, most or almost all the spatial map (e.g. detection, segmentation, depth, density) generation architectures \cite{faster-rcnn, mask-rcnn, deeplab-v3, panoptic, dense-depth, dorn, psmnet} employ the pretrained, computation-heavy encoders with comparatively much smaller decoders to transform the encoded features into the desired spatial maps. Therefore, placing an extra decoder for 2D segmentation parallel to the decoder in the monocular depth estimation model adds little overhead to the original monocular depth estimation. We train this auxiliary decoder with the shared feature maps from the encoder trained for monocular depth estimation, which we call depth-shared features.

This depth-shared features pretrained with mono depth estimation (MDE) datasets turns out to be useful for better 2D foreground extraction with comparatively smaller datasets which will be evident from ablation studies later. Note that the depth-shared decoder is trained conservatively insofar it has a sufficiently high recall with pragmatically good precision to not miss any object out of the blue in the preprocessing step. We argue that the 3D detector has the capacity to regress the correct bounding boxes from the softly refined PseudoLiDAR.

\section{Experiments}

\textbf{Datasets:} Evaluation is performed on the KITTI 3D object detection benchmark \cite{kitti3d} which is the \emph{de facto} standard for autonomous driving. This dataset comprises 7481 and 7518 images for training and leaderboard inference, respectively. 
The training set is further split into 3712 samples for training and 3769 for validation \cite{mono3d-2016} from the disjoint set of sequences to compare the proposed variants.
Moreover, based upon the perceived difficulty of detection, each ground truth instance is pigeonholed into one of the levels from \emph{\{``easy", ``moderate", ``hard"\} }.

\textbf{Rectified KITTI Metric:} Predicted bounding boxes are compared with ground truth using the \emph{N-point interpolated average precision (IAP)} metric \cite{kitti3d, interp-ap} as follows:

\vspace{-2ex}
\begin{equation}
AP|_{R_{N}} = \frac{1}{N} \sum_{r \in R_N} \underset{\tilde{r} \geq r}{\mathrm{max} } \; \rho(\tilde{r} ) 
\label{eq:metrics}
\end{equation}

\vspace{-1ex}
Until recently, \emph{11-point IAP} was employed in the official KITTI leaderboard with $R_N = \{0, 0.1, ..., 1\}$. 
However, the inclusion of $0$ caused the inflation of average precision by $\sim 9\%$ even with a tangential match \cite{disentangle_kitti}.
To avoid such ostensible boost in performance in a simple and elengant manner, both the metric and the leaderboard are revised with a new \emph{40-point IAP} (AP$|_{R_{40} }$) with the exclusion of ``0" and four-times denser interpolated prediction for better approximation of the area under the Precision/Recall curve. 

Thus, in this paper, we provide all the comparisons with the newly proposed AP$|_{R_{40} }$ metric, and completely jettison the old one (AP$|_{R_{11} }$) to avoid arguably inappropriate impression of the individual approaches. Note that almost all the previous works were both validated and tested on AP$|_{R_{11} }$. 
To accommodate this shift in terms of the comparison with recent literature, we benchmark our proposed variants using AP$|_{R_{40} }$ on the validation set, and 
compare with the updated \emph{APs} directly from the KITTI leaderboard for the test set that do not match the reported ones in the literature.

\textbf{Training and Implementation:} For monocular depth estimation (MDE), we train a DenseDepth \cite{dense-depth} variant on KITTI depth dataset \cite{kitti-mono-dataset}. The depth-shared decoder sub-network is trained with the frozen MDE encoder, thus safeguarding the depth estimator. Both the NN clustering in PoIS and AFgS modules are realized with ball query based on K-d tree \cite{ball-tree, kd-tree}.
PointRCNN \cite{point-rcnn} is chosen as our 3D detector as already mentioned in Section \ref{sec:approach}. The choice of optimization algorithms and hyperparameters are unaltered from the original.

\begin{table}[]
\centering
\begin{adjustbox}{width=0.42\textwidth}
\begin{tabular}{l|ccc}
\hline
\multicolumn{1}{c|}{\multirow{2}{*}{Method}} & \multicolumn{3}{c}{BEV / 3D} \\
\multicolumn{1}{c|}{} & Easy & Moderate & Hard \\ \hline
Baseline & 55.07/37.38 & 37.42/24.10 & 31.94/20.58 \\
Unsupervised & 55.95/41.31 & \textbf{38.08}/27.18 & \textbf{32.35}/22.42 \\
Supervised & \textbf{57.63/42.83} & 37.43/\textbf{27.73} & 31.63/\textbf{23.00} \\ \hline
\end{tabular}
\end{adjustbox}
\vspace{-1ex}
\caption{AP$|_{R_{40} }$ scores on KITTI3D \emph{``Car"} validation set.}
\label{tab:self_compare}
\end{table}

\begin{table}[t!]
\centering
\begin{adjustbox}{width=0.42\textwidth}
\begin{tabular}{l|ccc}
\hline
\multicolumn{1}{c|}{\multirow{2}{*}{Method}} & \multicolumn{3}{c}{BEV / 3D (IoU $\geq$ 0.7)} \\
 & Easy & Moderate & Hard \\ \hline
OFTNet \cite{oft-net} & 7.16/1.61 & 5.69/1.32 & 4.61/1.00 \\
FQNet \cite{fitting-degree} & 5.40/2.77 & 3.23/1.51 & 2.46/1.01 \\
ROI-10D \cite{roi-10d} & 9.78/4.32 & 4.91/2.02 & 3.74/1.46 \\
GS3D \cite{gs3d} & 8.41/4.47 & 6.08/2.90 & 4.94/2.47 \\
Shift-RCNN \cite{shift-rcnn} & 11.84/6.88 & 6.82/3.87 & 5.27/2.83 \\
MonoFENet \cite{mono-fenet} & 17.03/8.35 & 11.03/5.14 & 9.05/4.10 \\
MonoGRNet \cite{mono-grnet} & 18.19/9.61 & 11.17/5.74 & 8.73/4.25 \\
MonoPSR \cite{mono-psr} & 18.33/10.76 & 12.58/7.25 & 9.91/5.85 \\
MonoPL \cite{mono3d-pl} & 21.27/10.76 & 13.92/7.50 & 11.25/6.10 \\
SS3D \cite{mono-ss3d} & 16.33/10.78 & 11.52/7.68 & 9.93/6.51 \\
MonoDIS \cite{disentangle_kitti} & 17.23/10.37 & 13.19/7.94 & 11.12/6.40 \\
M3D-RPN \cite{m3d-rpn} & 21.02/14.76 & 13.67/9.71 & 10.23/7.42 \\
AM3D \cite{pseudolidar-copy} & 25.03/16.50 & 17.32/10.74 & \textbf{14.91/9.52} \\
\textcolor{blue!90!black}{RMPL (Ours)} & \textbf{28.08/18.09} & \textbf{17.60/11.14} & 13.95/8.94 \\ \hline
\end{tabular}
\end{adjustbox}
\vspace{-1ex}
\caption{AP$|_{R_{40} }$ scores on KITTI3D \emph{``Car"} test set.}
\label{tab:car_kitti}
\end{table}

\begin{table}[ht!]
\centering
\begin{adjustbox}{width=0.42\textwidth}
\begin{tabular}{l|ccc}
\hline
\multicolumn{1}{c|}{\multirow{2}{*}{Method}} & \multicolumn{3}{c}{BEV / 3D (IoU $\geq$ 0.5)} \\
 & Easy & Moderate & Hard \\ \hline
OFTNet \cite{oft-net} & 1.28/0.63 & 0.81/0.36 & 0.51/0.35 \\
SS3D \cite{mono-ss3d} & 2.48/2.31 & 2.09/1.78 & 1.61/1.48 \\
M3D-RPN \cite{m3d-rpn} & 5.65/4.92 & 4.05/3.48 & 3.29/2.94 \\
MonoPSR \cite{mono-psr} & 7.24/6.12 & 4.56/4.00 & 4.11/3.30 \\
Shift-RCNN \cite{shift-rcnn} & 8.58/7.95 & 5.66/4.66 & 4.49/4.16 \\
\textcolor{blue!90!black}{RMPL (Ours)} & \textbf{13.09/11.14} & \textbf{7.92/7.18} & \textbf{7.25/5.84} \\ \hline
\end{tabular}
\end{adjustbox}
\vspace{-1ex}
\caption{AP$|_{R_{40} }$ scores on KITTI3D \emph{``Pedestrian"} test set.}
\label{tab:ped_kitti}
\end{table}

\begin{figure*}[!ht]
\centering
\hspace{0ex}
\includegraphics[width=0.49\textwidth,angle=0]{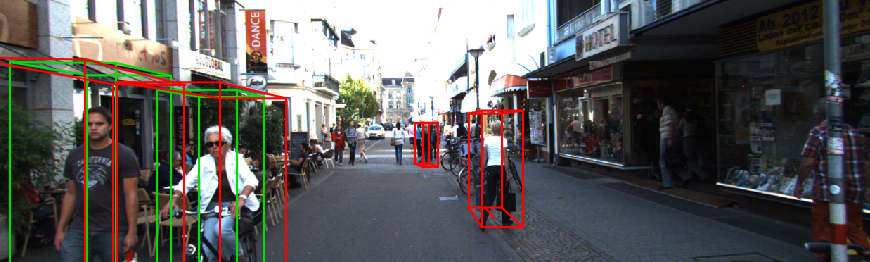}
\hspace{0ex}
\includegraphics[width=0.49\textwidth,angle=0]{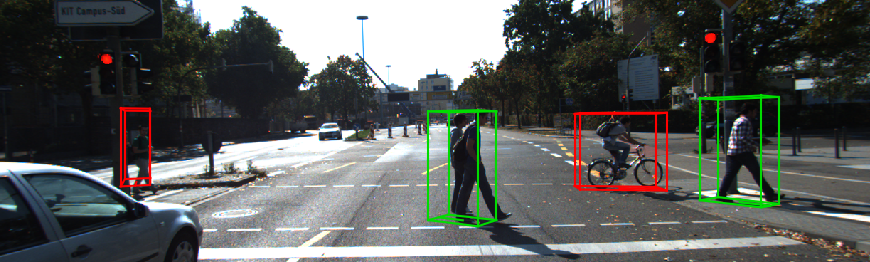}
\vspace{-1.5ex}
\caption{Intermingled predictions for \emph{pedestrian} (green) and \emph{cyclist} (red) categories. High spatial similarities between these two classes due to the common constituent (humans) make their detection problems inextricably interwoven.} 
\label{fig:confusion}
\end{figure*}

\subsection{Results}

\textbf{Baseline comparison:} Table \ref{tab:self_compare} lists BEV and 3D APs for the baseline and both of our unsupervised and supervised schemes for the KITTI3D ``Car" validation set. Note that we do not include recent methods here for comparison since they provide arguably deceptive AP$|_{R_{11} }$ scores (including $0$-Recall) only. Both of our data engineering approaches perform better than the baseline counterpart. From this table, it is evident that unlike 2D detectors which do not require \emph{a priori} segmentation for accurate bounding box regression \cite{faster-rcnn, yolo9000, ssd}, 3D detectors cannot extract sufficient information from the highly redundant point clouds like PseudoLiDARs themselves. As a result, further refinement or segmentation of point clouds prior to detection assists in performance improvement here. Also, Table \ref{tab:self_compare} fails to discern the absolute winner between our two policies. Nonetheless, due to its slight numerical gain, supervised schema is chosen for the KITTI leaderboard.

\textbf{Leaderboard comparison:} Table \ref{tab:car_kitti}, \ref{tab:ped_kitti}, and \ref{tab:cyc_kitti} compare our method (RMPL) using AP$|_{R_{40} }$ scores on the KITTI 3D and BEV leaderboard with recent literature for ``Car", ``Pedestrian", and ``Cyclist" categories, respectively.

For \emph{``Car"}, we achieve SOTA results on both \emph{easy} and \emph{moderate} sub-categories while lagging behind AM3D \cite{disentangle_kitti} on the \emph{hard} cases. 
We argue that the comparatively poorer performance on the \emph{hard} instances are mostly due to the ill-posed nature of the mono depth esimation problem. Note that MDE is a core component in our framework that is not the case for AM3D \cite{disentangle_kitti}. More light on this issue will be cast in Section \ref{sec:failure} later.  

For \emph{``Pedestrian"}, from all aspects, our method outplays others by a large margin with $54\%$ relative improvement for 3D \emph{moderate} AP in particular. This significant gain should be credited to the constructive reduction in ambiguity by our refinement strategies for smaller objects like pedestrians.

For \emph{``Cyclist"}, we stand second behind MonoPSR \cite{mono-psr} with a non-negligible margin ($1.82$ \emph{vs.} $4.74$ in Table \ref{tab:cyc_kitti}). Note that we excel MonoPSR \cite{mono-psr} by a similar margin ($7.18$ \emph{vs.} $4.00$ in Table \ref{tab:ped_kitti}) on the \emph{``Pedestrian"} leaderboard. The reason behind this shift in domination can be hypothesized to be an internal shift in the parameter space of the responsible models. Both the 2D and 3D detection models are affected by the contextual and structural similarities of \emph{pedestrian} and \emph{cyclist} classes because cyclist instances can be modeled as a soft set \cite{soft-set} under the universe comprising parts of a human and a bicycle in the spatial domain. This is also partially indicated by the poorer performance of LiDAR-based systems on these two categories compared to that of cars \cite{pointpillars, point-rcnn}. Figure \ref{fig:confusion} shows intermingled detections for \emph{pedestrian} and \emph{cyclist} on KITTI validation images. Despite the good quality of detection, such arguably inevitable confusions leads to the severe degradation of our \emph{cyclist} category, and possibly the \emph{pedestrian} as well. 

\begin{figure*}[!ht]
\centering
\includegraphics[width=0.482\textwidth,angle=0]{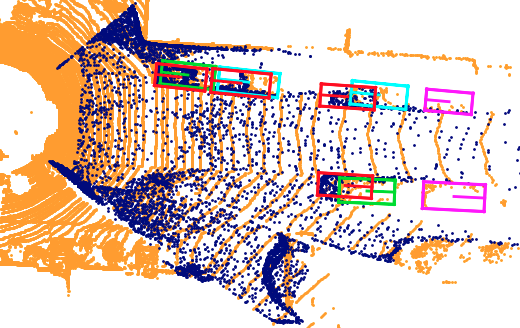}
\hspace{2ex}
\includegraphics[width=0.482\textwidth,angle=0]{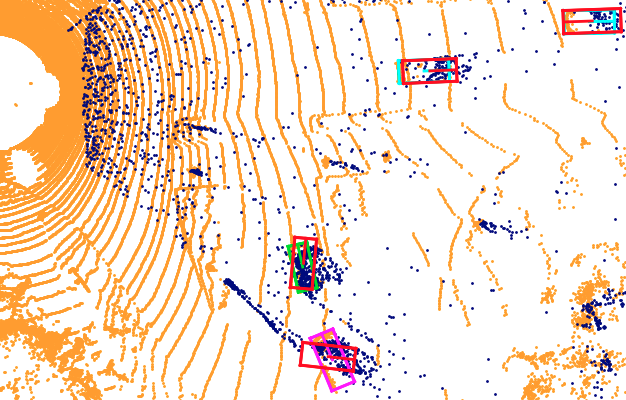}  \\
\hspace{-1.1ex}
\includegraphics[width=0.482\textwidth,angle=0]{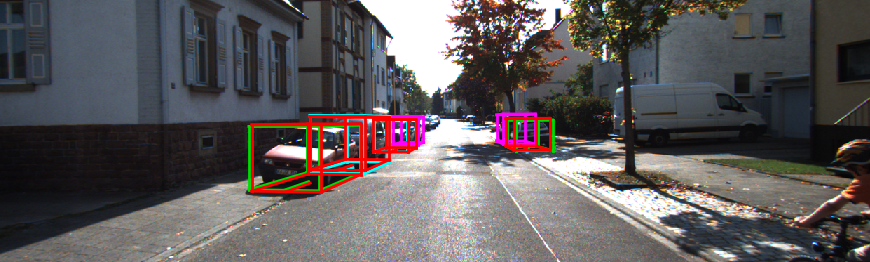}
\hspace{2ex}
\includegraphics[width=0.482\textwidth,angle=0]{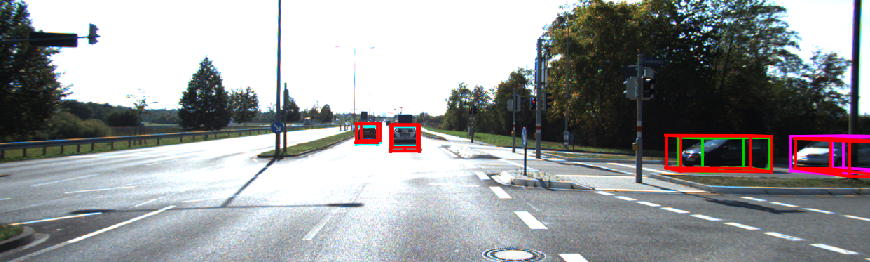} \\
\includegraphics[width=0.482\textwidth,angle=0]{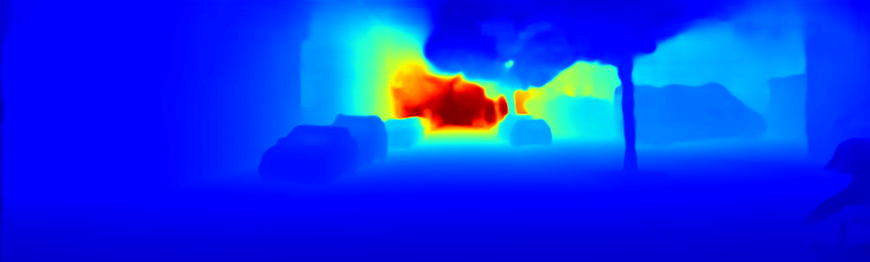} 
\hspace{2ex}
\includegraphics[width=0.482\textwidth,angle=0]{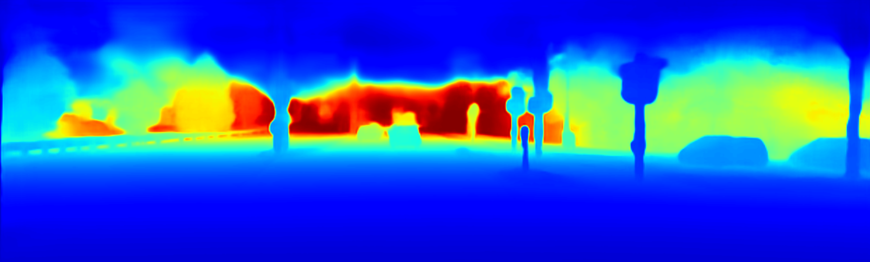}
\vspace{-1ex}
\caption{Cropped BEV (top) and RGB (middle) with superimposed predictions \emph{(red)} and ground truth \emph{(easy--green, moderate--cyan, hard--magenta)}, and depth prediction (bottom). BEV LiDAR and sparse PseudoLiDAR are shown in \emph{orange} and \emph{dark blue}, respectively. Left sample exemplifies the nonlinear shift of far away instances due to ill-posed mono depth prediction. Right sample exhibits erroneous estimation of object orientation due to its rotation.}
\label{fig:misalignment}
\end{figure*}

\subsection{Ablation Studies}
\label{sec:ablation}

Experimental ablation is done on two aspects of our pipeline -- (1) Effect of the sampling rate of DSS module (Table \ref{tab:dss}), and (2) Effect of depth-shared encoder for 2D detection (Table \ref{tab:dsd}).

Table \ref{tab:dss} indeed manifests that the extravagant points in the raw PseudoLiDAR restrain the 3D detector to achieve an overall high performance than assisting it. As shown in Figure \ref{fig:starter}, the stratified reduction in the density of points by DSS helps to differentiate between the sporadically high density background point clusters and foreground ones by conserving the spatial enclosure of foreground instances.

Table \ref{tab:dsd} provides comparison between the detection APs for our depth-shared detector (DSD) and the same detector trained with a dedicated encoder pretrained on ImageNet \cite{imagenet}. Our DSD precision is on a par with that of separate encoder. However, given a much higher computational complexity of the encoder compared to the decoder in the standard models \cite{faster-rcnn, mask-rcnn, deeplab-v3}, we claim that DSD adds very little computational overhead on MDE (essential for PseudoLiDAR generation) than a separate detector network.

\begin{table}[th!]
\centering
\begin{adjustbox}{width=0.42\textwidth}
\begin{tabular}{l|ccc}
\hline
\multicolumn{1}{c|}{\multirow{2}{*}{Method}} & \multicolumn{3}{c}{BEV / 3D (IoU $\geq$ 0.5)} \\
 & Easy & Moderate & Hard \\ \hline
OFTNet \cite{oft-net} & 0.36/0.14 & 0.16/0.06 & 0.15/0.07 \\
Shift-RCNN \cite{shift-rcnn} & 0.76/0.48 & 0.38/0.29 & 0.41/0.31 \\
M3D-RPN \cite{m3d-rpn} & 1.25/0.94 & 0.81/0.65 & 0.78/0.47 \\
SS3D \cite{mono-ss3d} & 3.45/2.80 & 1.89/1.45 & 1.44/1.35 \\
MonoPSR \cite{mono-psr} & \textbf{9.87/8.37} & \textbf{5.78/4.74} & \textbf{4.57/3.68} \\
\textcolor{blue!90!black}{RMPL (Ours)} & 4.23/3.23 & 2.42/1.82 & 2.14/1.77 \\ \hline
\end{tabular}
\end{adjustbox}
\vspace{-1ex}
\caption{AP$|_{R_{40} }$ scores on KITTI3D \emph{``Cyclist"} test set.}
\label{tab:cyc_kitti}
\end{table}

\vspace{-2.5ex}

\begin{table}[!ht]
\centering
\begin{adjustbox}{width=0.45\textwidth}
\begin{tabular}{l|ccc}
\hline
\multicolumn{1}{c|}{\multirow{2}{*}{Method}} & \multicolumn{3}{c}{BEV / 3D (IoU $\geq$ 0.7)} \\
 & Easy & Moderate & Hard \\ \hline
B (no DSS)  & 53.91/35.92 & 35.70/22.41 & 30.43/19.90 \\ 
B-DSS(80\%) & 53.54/36.36 & 35.34/22.33 & 29.86/19.53 \\
B-DSS(60\%) & 54.17/34.53 & 36.08/21.93 & 30.64/18.72 \\
B-DSS(40\%) & 52.15/35.48 & 35.25/23.33 & 29.93/20.05 \\
B-DSS(20\%) & 53.19/34.23 & 35.29/22.35 & 30.01/19.34 \\
B-DSS(10\%) & 55.07/37.38 & 37.42/24.10 & 31.94/20.58 \\ \hline
\end{tabular}
\end{adjustbox}
\vspace{-1ex}
\caption{Effect of DSS sampling rate on AP$|_{R_{40} }$ scores for KITTI3D \emph{``Car"} validation set. \emph{B $\equiv$ Baseline} }
\label{tab:dss}
\end{table}

\vspace{-2.5ex}

\begin{table}[!ht]
\centering
\begin{adjustbox}{width=0.48\textwidth}
\begin{tabular}{l|ccc}
\hline
\multicolumn{1}{c|}{\multirow{2}{*}{Method}} & \multicolumn{3}{c}{BEV / 3D (IoU $\geq$ 0.7)} \\
 & Easy & Moderate & Hard \\ \hline
S-Detector(separate) & 56.32/40.65 & \textbf{37.50}/27.66 & \textbf{32.03/23.20} \\
S-DSD \textcolor{blue!90!black}{(RMPL)} & \textbf{57.63/42.83} & 37.43/\textbf{27.73} & 31.63/23.00 \\ \hline
\end{tabular}
\end{adjustbox}
\vspace{-1ex}
\caption{Effect of DSD on AP$|_{R_{40} }$ scores for KITTI3D \emph{``Car"} validation set. \emph{S $\equiv$ Supervised} }
\label{tab:dsd}
\end{table}

\begin{table}[!ht]
\centering
\begin{adjustbox}{width=0.45\textwidth}
\begin{tabular}{c|ccc}
\hline
\multicolumn{1}{c|}{\multirow{2}{*}{Range (meters)}} & \multicolumn{3}{c}{BEV / 3D (IoU $\geq$ 0.7)} \\
 & Easy & Moderate & Hard \\ \hline
$0-10$  & 73.36/52.66 & 79.33/61.17 & 80.60/61.72 \\ 
$0-20$  & 70.51/54.85 & 66.93/53.30 & 58.36/46.03 \\ 
$0-30$  & 59.75/44.50 & 53.62/40.40 & 43.60/33.14 \\ 
$0-40$  & 57.66/42.83 & 42.19/31.69 & 34.53/25.34 \\ 
$0-50$  & 57.63/42.83 & 37.51/27.78 & 31.69/23.04 \\ 
$0-60$  & 57.63/42.83 & 37.43/27.73 & 31.63/23.00 \\ 
Full & 57.63/42.83 & 37.43/27.73 & 31.63/23.00 \\ \hline
\end{tabular}
\end{adjustbox}
\vspace{-1ex}
\caption{Effect of instance-depth on AP$|_{R_{40} }$ scores for KITTI3D \emph{``Car"} validation set. }
\label{tab:depth_wise_eval}
\end{table}

\vspace{-2ex}
\subsection{Failure Analysis}
\label{sec:failure}

In this section, we attempt to characterise the possible failure cases causing a significantly large gap between our monocular 3D detection system and LiDAR-based frameworks. Table \ref{tab:depth_wise_eval} lists AP$|_{R_{40} }$ of our supervised pipeline for the ground truth sets with the gradually increased range of depth. Although not quite comparable to the \emph{moderate} AP of LiDAR-based PointRCNN ($75.64\%$), the accuracy doubles ($53.30\%$ \emph{vs.} $27.73\%$) for instances located close to the ego vehicle ($0-20$ meters). The reason for this acute degradation is manifold. First, the quality of mono depth prediction drops drastically with the increase of depth \cite{bts, dorn}, which causes an instance-wise, nonlinear shift of the points (Figure \ref{fig:misalignment} left image). This nonlinear, pointwise translation eventually offsets the 3D bounding boxes regressed by the detectors compared to the corresponding ground truth boxes. Second, although the prediction of MDE appears both accurate and appealing, it fails to recognise the direction of the slanted or rotated objects, thus resulting in an unavoidable 3D prediction error (Figure \ref{fig:misalignment} right image). Finally, far away objects are more likely to be occluded, which could be another major cause of deterioration of 3D APs with the increase in depth.

\section{Conclusion}

In this paper, we explore the data engineering approach to improve the quality of PseudoLiDAR point clouds for 3D detection. First, we show the significant redundancy and perplexing nature of the raw PseudoLiDAR point clouds for 3D detection due to its extraordinarily high density. Next, we endeavour to overcome this ambiguity with a couple of strategies for sparsification without overburdening from the computational perspective. We obtain SOTA performance on \emph{``Car"} and \emph{``Pedestrian"} categories on the KITTI leaderboard. Our model (detector) and process (mono/stereo) agnostic sparsifications demonstrate the necessity of data orchestration alongside architectural enhancement, especially for generated data like PseudoLiDAR. Finally, we elucidate the root causes behind the inconsistency of monocular performance with LiDAR counterparts, which could be good directions for future exploration.

{\small
\bibliographystyle{ieee} 
\bibliography{3d_detection,mono_3d,bev_3d,stereo,mono_depth,stereo_depth,books,general}
}

\end{document}